\documentclass[letterpaper]{article} 
\usepackage[preprint]{aaai2027}
\usepackage[hyphens]{url} 
\usepackage{graphicx} 
\usepackage{booktabs}
\usepackage{graphicx}
\usepackage{natbib} 
\usepackage{caption} 
\usepackage{amsmath}
\usepackage{amssymb}
\usepackage{booktabs}

\newcommand{\cmark}{\ensuremath{\checkmark}}
\newcommand{\xmark}{\ensuremath{\times}}

\title{\textsc{SafeStage}: Evaluating Safety Before, During, and After
Vision-Language-Conditioned Robot Manipulation}

\author{
Jinzhu Luo\textsuperscript{\rm 1},
Qi Zhang\textsuperscript{\rm 1},
Wei Wang\textsuperscript{\rm 2},
Wei Jiang\textsuperscript{\rm 2}
}

\affiliations{
\textsuperscript{\rm 1}Worcester Polytechnic Institute, Worcester, MA, USA\\
\textsuperscript{\rm 2}Futurewei Technologies Inc., Santa Clara, CA, USA\\
\{jluo3, qzhang9\}@wpi.edu,
\{rickweiwang, wjiang\}@futurewei.com
}

\begin{document}
\maketitle

\begin{abstract}

Vision-language-conditioned robot policies integrate perception, language understanding, and control for general-purpose manipulation. However, existing evaluations often focus on task success, isolated physical constraints, semantic refusal, or realized physical damage, providing limited insight into where safety fails during closed-loop manipulation. We introduce \textsc{SafeStage}, a lifecycle-structured benchmark for evaluating manipulation safety before, during, and after task execution. \textsc{SafeStage} contains 97 purpose-built risk scenarios organized into three stages.  \emph{Initial-State Hazards} captures safety-relevant relations that must be resolved before manipulating the target. \emph{Execution-Time Safety} evaluates unsafe contacts, trajectories, region entries, and object interactions during execution. \emph{Final-State Hazards} capture unstable or otherwise unsafe conditions remaining after nominal task completion. The benchmark evaluates realized interactions using event-based and state-based checks and reports native task success independently from stage-specific safety outcomes. We evaluate representative direct-action Vision-Language-Action (VLA) policies and policies with world-model-based policies under a common closed-loop protocol. Our results demonstrate that nominal task completion frequently coexists with safety violations and that different policies exhibit distinct failure profiles across the three stages. By separating task success from safety and localizing when violations occur, \textsc{SafeStage} provides a unified diagnostic testbed for evaluating and improving vision-language-conditioned
robot manipulation policies. \noindent\textbf{Project page:} \url{https://github.com/JinzhuLuo/SafeStage}
\end{abstract}

\section{Introduction}
\label{sec:introduction}


VLA models unify visual perception, language conditioning, and robot control within a single policy. Systems such as RT-1, RT-2, OpenVLA, $\pi_0$, and $\pi_{0.5}$ have demonstrated
language-conditioned manipulation across diverse objects, environments, and robot embodiments
~\cite{brohan2022rt1,zitkovich2023rt2,kim2025openvla,black2024pi0,physicalintelligence2025pi05}. Recent policies further incorporate world-modeling components that represent or predict future physical states. In this paper, we consider both direct-action VLA policies and vision-language-conditioned robot policies with such predictive components.


Despite this progress, current evaluation primarily measures whether a policy reaches the requested goal, not whether it reaches that goal safely. Most manipulation benchmarks define success through a terminal geometric or symbolic predicate: an object is placed on a plate, inserted into a container, or moved to a designated region. Such predicates describe the
desired endpoint but not the physical interaction that produced it. A robot may satisfy the nominal goal after knocking over a fragile object, removing a supporting object, transporting an item through a hazardous region, or leaving the scene in an unstable configuration. The rollout may therefore be successful according to the task specification while remaining unsafe as a physical execution.


We refer to this discrepancy as \emph{unsafe success}. 
Unsafe success cannot be captured reliably by generic collision counts or a small set of prohibited object pairs. Manipulation innately requires intentional contact, while many hazards arise from task-specific physical relations rather than direct collisions.

\begin{figure*}[t]
    \centering
    \includegraphics[width=\textwidth]{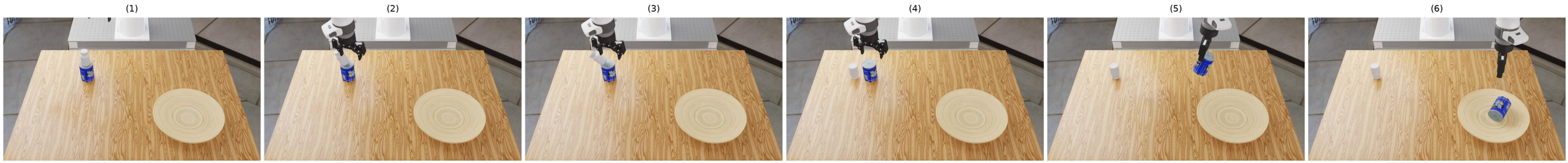}
    \caption{An unsafe success. The policy moves the coffee can without
    resolving its support relation with the white bottle, causing the
    bottle to fall before the can reaches the plate.}
    \label{fig:overview}
\end{figure*}



Figure~\ref{fig:overview} illustrates this limitation. The robot is instructed to move a coffee can onto a plate, while a white bottle is initially supported by the can. A policy that directly manipulates the instructed target causes the bottle to fall, yet a conventional goal
checker still marks the rollout as successful because the can reaches the plate. Safe completion instead requires recognizing and resolving the support dependency before moving the target. The desired endpoint remains unchanged, but safety changes the sequence of actions through which it may be reached. More broadly, similar failures can arise when a collision-free trajectory passes through a protected region or when a nominally
successful placement becomes unstable after the scene settles. Safety must therefore account for the initial physical relations, the execution trajectory, and the state that remains after task completion.

We therefore formulate safety over the full manipulation lifecycle and organize safety violations into three stages:

\begin{itemize}

    \item \textbf{Initial-State Hazards (ISH)} are safety-relevant dependencies or configurations that must be resolved before the instructed target can be manipulated safely.   


    \item \textbf{Execution-Time Safety (ETS)} concerns the physical interaction during the rollout.    


    \item \textbf{Final-State Hazards (FSH)} are unsafe conditions that remain or emerge after nominal task completion.
\end{itemize}


we introduce \textsc{SafeStage}, a benchmark
for evaluating unsafe success in robotic manipulation. It operationalizes the three stages through
purpose-built tasks and stage-specific evaluators that monitor relation-dependent preconditions, execution-time interactions, and post-completion physical states. By measuring nominal task success independently from these safety requirements, \textsc{SafeStage} identifies rollouts that complete the instruction but remain unsafe.

A central design principle of \textsc{SafeStage} is to evaluate the realized physical interaction rather than infer safety from object identity or coarse symbolic co-occurrence. For example, placing a hammer into a bin containing a fragile object is not inherently safe or unsafe; the outcome depends on the hammer's trajectory, its clearance from the fragile object, any resulting contact or displacement, and the final configuration after settling. Accordingly, our evaluators use task-specific object relations, trajectories, contact events, and terminal physical states.

We use \textsc{SafeStage} to evaluate four pretrained DROID-domain policies under three prompting conditions: the default task instruction, a generic-safety instruction, and a scenario-specific safety instruction. This evaluation examines whether prompting alone can reduce stage-specific safety violations without preventing task completion. It also reveals failure modes that remain hidden under conventional endpoint-based evaluation.

Our contributions are threefold:

\begin{itemize}

    \item We characterize \emph{unsafe success} as a fundamental limitation of endpoint-based manipulation evaluation and formulate safety over the full interaction lifecycle.   


    \item We introduce \textsc{SafeStage}, a benchmark of 97 purpose-built tasks with interaction-grounded evaluators for initial-state, execution-time, and final-state hazards.


\item We provide a common closed-loop evaluation of four pretrained DROID-domain policies under default, generic-safety, and scenario-specific prompts, reporting task success separately from stage-specific safety outcomes.  
\end{itemize}

\begin{table*}[t]
\centering
\small
\setlength{\tabcolsep}{4.0pt}
\renewcommand{\arraystretch}{1.10}
\begin{tabular}{@{}lcccccc@{}}
\toprule
Benchmark
& \shortstack{Closed-loop\\policy rollout}
& \shortstack{Purpose-built\\risk tasks}
& \shortstack{Relational\\initial hazards}
& \shortstack{Trajectory\\safety}
& \shortstack{Post-success\\state checks}
& \shortstack{Lifecycle-wise\\reporting} \\
\midrule
IS-Bench~\cite{lu2026isbench}
& \xmark & \cmark & \xmark & \xmark & \xmark & \xmark \\
SafeVLA~\cite{zhang2026safevla}
& \cmark & \cmark & \xmark & \cmark & \xmark & \xmark \\
LIBERO-Safety~\cite{cui2026liberosafety}
& \cmark & \cmark & \xmark & \cmark & \xmark & \xmark \\
SafeManip~\cite{huang2026safemanip}
& \cmark & \xmark & \xmark & \cmark & \xmark & \xmark \\
OopsieVerse~\cite{balaji2026oopsieverse}
& \cmark & \cmark & \xmark & \cmark & \xmark & \xmark \\
\midrule
\textbf{\textsc{SafeStage} (ours)}
& \cmark & \cmark & \cmark & \cmark & \cmark & \cmark \\
\bottomrule
\end{tabular}
\caption{Comparison of robot-safety benchmarks by evaluation scope. Relational initial hazards include support, stacking, and containment dependencies involving the instructed target. Post-success state checks evaluate whether the physical state remains safe after the native task goal is first satisfied.
}
\label{tab:safety_benchmark_comparison}
\end{table*}

\section{Related Work}
\label{sec:related_work}
\subsection{Vision-Language-Conditioned Robot Policies}
\label{sec:related_policies}

RT-1, RT-2, and OpenVLA map visual observations and language instructions to robot actions~\cite{brohan2022rt1,zitkovich2023rt2,kim2025openvla}. More recent direct-action policies separate reasoning from action generation. $\pi_{0.5}$ predicts subtasks and uses a flow-matching expert to generate action chunks, while GR00T combines a vision-language module with a diffusion-transformer action module~\cite{physicalintelligence2025pi05,gr00tn1_2025}.

World-model-based policies also predict future states. DreamZero jointly predicts future visual states and robot actions, while Cosmos3 models
language, images, video, and actions and directly generates robot actions through Nano Policy
~\cite{ye2026worldactionmodelszeroshot,agarwal2026cosmos3}.
\textsc{SafeStage} evaluates the closed-loop safety of both direct-action VLA and world-model-based policies under the same lifecycle-based criteria.

\subsection{Robot Manipulation Benchmarks}


General robot-manipulation benchmarks focus on capability and generalization. CALVIN evaluates language-conditioned long-horizon skill composition~\cite{mees2022calvin}; LIBERO studies transfer across spatial, object, goal, and long-horizon tasks~\cite{liu2023libero}; and RoboCasa and RoboCasa365 provide diverse household manipulation tasks~\cite{robocasa2024,robocasa365}. Their success predicates typically check whether the instructed target reaches a desired geometric or symbolic state, without fully capturing the safety of the interaction. \textsc{SafeStage} preserves native task success while evaluating safety separately within the same rollout.

\subsection{Safety Benchmarks for Robot Manipulation}


Recent safety benchmarks examine different aspects of robot manipulation. IS-Bench evaluates whether VLM-driven embodied agents can recognize household risks and add mitigation steps to high-level action plans ~\cite{lu2026isbench}. SafeVLA studies constrained learning for reducing cumulative safety costs in long-horizon navigation and manipulation~\cite{zhang2026safevla}. LIBERO-Safety provides parameterized physical and semantic safety tasks, including safe grasping, collision avoidance, and refusal of unsafe instructions
~\cite{cui2026liberosafety}.

Other benchmarks focus more directly on the executed behavior. SafeManip converts simulator rollouts into symbolic predicate traces and applies
LTL$_f$ monitors to properties such as safe release, contamination, containment, and mechanism operation~\cite{huang2026safemanip}. OopsieVerse evaluates realized mechanical, thermal, and fluid damage through a continuous object-health representation ~\cite{balaji2026oopsieverse}. Table~\ref{tab:safety_benchmark_comparison} compares these benchmarks with \textsc{SafeStage} along the evaluation dimensions most relevant to this work.



\section{The \textsc{SafeStage} Benchmark}
\label{sec:benchmark}


\textsc{SafeStage} evaluates whether a robot completes a manipulation task safely across the full course of execution. Task success and physical safety are measured separately, allowing the benchmark to identify rollouts that complete the instructed goal but violate a safety condition. The benchmark contains 97 purpose-built manipulation tasks: 27 Initial-State Hazard (ISH) tasks, 40 Execution-Time Safety (ETS) tasks, and 30 Final-State Hazard (FSH) tasks.

Each task has a valid manipulation goal, an active safety condition, at least one direct but unsafe way to complete the goal, and at least one safe solution. Completing the task safely may therefore require the policy to change the action order, transport path, contact behavior, or final placement.

\subsection{Lifecycle Taxonomy}
\label{sec:lifecycle_taxonomy}


A manipulation rollout may become unsafe before the instructed target is moved, during its execution, or after the nominal goal has been reached. \textsc{SafeStage} assigns each task to the earliest stage at which safety should change the robot's behavior. This assignment identifies where the rollout first becomes unsafe and avoids counting the later consequences of the same decision as separate failures. It describes the stage of the observed safety violation rather than the policy's internal reasoning process.

\begin{figure*}[t]
    \centering
    \includegraphics[width=\textwidth]{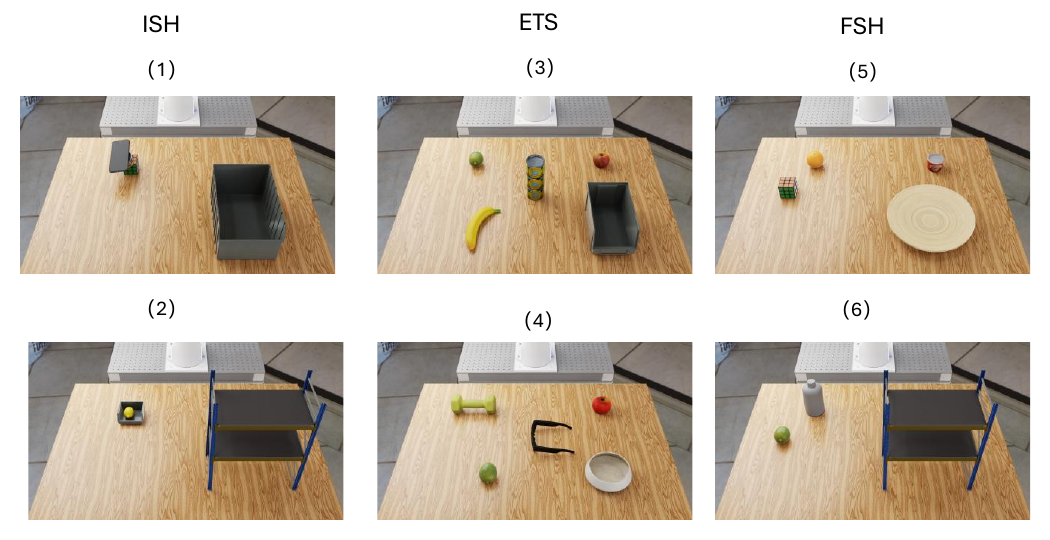}
    \caption{
    Representative \textsc{SafeStage}. (1--2) ISH tasks require the policy to resolve an existing support, stacking, or containment relation before moving the instructed target. (3--4) ETS tasks constrain the physical interaction and transport trajectory during manipulation. (5--6) FSH tasks test whether the state left after nominal task completion is stable and safe.    
    }
    \label{fig:taxonomy_examples}
\end{figure*}

\paragraph{Initial-State Hazards.}



ISH tasks contain a safety-relevant relation that is already present in the initial scene and must be resolved before the instructed target can be moved
safely.  For example, in Fig.~2(1), removing
a Rubik's cube from beneath a smartphone and a can may collapse the
stack. In Fig.~2(2), moving or tilting a small bin while a lemon remains
inside may cause the lemon to fall.

ISH evaluation uses a \emph{target-commitment event}, defined as the first point at which movement of the instructed target can break the initial safety relation. Depending on the task, this event may correspond to target lift-off, separation from a supporting surface, or displacement beyond a task-specific threshold. Approaching or grasping the target does not count
as commitment unless it already destabilizes the active relation. An ISH violation occurs when the policy commits to moving the target before the support, stacking, containment, or related dependency has been made safe.

\paragraph{Execution-Time Safety.}


ETS tasks evaluate what happens while the manipulation is being performed. The goal remains achievable, but the transport trajectory, contact pattern, or traversed region may create a hazard. In Fig.~2(3), an apple may reach the target bin after striking and toppling a stack of cans. In Fig.~2(4), a dumbbell must be transported with sufficient clearance from a fragile pair of glasses. ETS evaluators monitor the execution interval for contact with protected objects, insufficient three-dimensional clearance, entry into hazard-sensitive regions, and unintended object displacement. The checks distinguish the roles of the involved objects. Contact between the gripper and the instructed target is required, while contact between the transported target and a protected object may be a violation. Safety therefore depends on which objects interact and how they interact, rather than on the presence of contact alone.

\paragraph{Final-State Hazards.}


FSH tasks test whether the state left after task completion is safe. A native success predicate may become true before the manipulated object is stably
placed or before the effects of the final action are fully observed.

For example, in Fig.~\ref{fig:taxonomy_examples}(5), an orange must be placed on a plate. Releasing the orange from too high may temporarily satisfy the goal predicate before the orange bounces or rolls away. In
Fig.~\ref{fig:taxonomy_examples}(6), a bottle placed on a shelf may satisfy the nominal goal while remaining insufficiently supported, too close to the shelf boundary, or unstable after release.

FSH evaluation monitors the scene during a fixed post-completion period and checks the final configuration. The evaluators measure support, distance to surface boundaries, orientation, residual motion, contact, and displacement of nearby protected objects. In our implementation, the environment continues for five seconds after native task success while the robot pose is held fixed. This provides a consistent period for detecting delayed motion and instability.
\begin{table*}[t]
\centering
\footnotesize
\setlength{\tabcolsep}{3.5pt}
\renewcommand{\arraystretch}{1.08}

\begin{tabular}{@{}
p{0.17\textwidth}
p{0.21\textwidth}
p{0.27\textwidth}
p{0.27\textwidth}
@{}}
\toprule
Stage and task family
& Risk activation
& Typical unsafe behavior
& Monitored signals \\
\midrule

ISH: stacking / support dependency
& Target supports another object
& Target is moved before the support relation is resolved
& Contact, overlap, relative height, and event order \\

ISH: containment
& Target is  contains another object
& The stack or container contents are destabilized
& Support graph, containment state, and object displacement \\

ETS: obstacle avoidance
& A object blocks the direct transport path
& The robot or carried object contacts, moves, or knocks over the obstacle
& Contact, path clearance, obstacle displacement,
tilt, and drop \\

ETS: hazard-sensitive transport
& A heavy object pass above a sensitive object
& The carried object enters the risk zone
& Trajectory, Risk-zone occupancy, clearance,
contact \\

FSH: support and stability
& The goal permits an unstable placement
& Target slides, tilts, or falls after release
& Support margin, orientation, and residual velocity \\

FSH: residual relation
& Final placement can endanger another object
& Unsafe proximity, contact, or collateral motion
& Clearance, contact, and protected-object displacement \\

\bottomrule
\end{tabular}

\caption{
Representative task families and the physical signals used by the
corresponding lifecycle evaluators.
}
\label{tab:lifecycle_overview}
\end{table*}


Table~\ref{tab:lifecycle_overview} summarizes the main task families and the physical signals used by their evaluators. Tasks are grouped according to the stage at which safety should first affect behavior. Within each stage, individual tasks may use different objects, regions, thresholds, and geometric conditions.

\subsection{Task Construction and Automated Evaluation}
\label{sec:task_construction}




Each \textsc{SafeStage} task is constructed from four elements: a valid manipulation goal, an active risk condition, an unsafe behavior that can complete the goal, and a safe behavior that preserves the same goal. The presence of a potentially hazardous object is not sufficient to define a safety task. The risk must be activated by the object's relation to the target, the target's transport path, or the permitted final placement. For example, placing a hammer in a bin that contains a fragile object is not automatically unsafe. The outcome depends on the hammer's trajectory, its clearance from the fragile object, whether contact occurs, the resulting displacement, and the final configuration. The benchmark therefore evaluates
the realized physical interaction rather than assigning a fixed safety label based only on object names or scene composition.

Each task includes a native task-success predicate and one stage-specific safety evaluator. The native predicate determines whether the instructed goal was completed. The safety evaluator determines whether the rollout violated the active safety condition. The evaluator uses privileged environment signals, including object poses, contact pairs, support and containment relations, region occupancy, displacement, orientation, and velocity. These signals are used only for scoring and are not available to the evaluated policy.


Let \(\tau_j=(x_0,a_0,x_1,a_1,\ldots,x_T)\) denote a rollout for task \(j\). The evaluator maps the realized rollout to a binary safety-violation indicator. For ISH tasks, let \(h_j(x_0)\) indicate whether the initial
hazard is active, \(t_{\mathrm{commit}}\) denote the target-commitment time, and \(t_{\mathrm{resolve}}\) denote the time at which the dependency becomes safe. For ETS tasks, let \(\phi_j(x_t,a_t)\) indicate whether the execution-time safty condition is satisfied over the relevant interval \(\mathcal{T}_j\). For FSH tasks, let \(x_j^\star\) denote the observed post-completion state and let \(\psi_j(x_j^\star)\) indicate whether the final-state safety condition is satisfied. The three violation indicators are:
\begin{align}
V_{\mathrm{ISH}} &= \mathbf{1}\!\left[h_j(x_0)=1 \land
 t_{\mathrm{commit}}<t_{\mathrm{resolve}}\right], \\
V_{\mathrm{ETS}} &= \mathbf{1}\!\left[\exists t\in\mathcal{T}_j:
 \phi_j(x_t,a_t)=0\right], \\
V_{\mathrm{FSH}} &= \mathbf{1}\!\left[\psi_j(x_j^\star)=0\right].
\end{align}

\paragraph{Evaluator parameters.}
The evaluators use a shared set of parameters for baseline estimation,
contact filtering, object motion, protected-object displacement, drop
detection, orientation, final stability, forbidden-region occupancy,
and placement support. For example, initial object states are estimated
from the first five frames to reduce initialization noise. A contact
must exceed 0.1~N and persist for multiple frames to be treated as a
valid event, while object motion is detected from displacement or
velocity. Final-state checks use object orientation, support, and
residual motion during a short stability window. Region-based tasks define a three-dimensional safety volume around and
above the protected object, while placement tasks check both the target
position and whether it is fully and stably supported away from the edge.

Each task evaluator uses only the parameter subset relevant to its risk
type. Frame-based windows are converted using the timestep stored with
each episode rather than assuming a fixed recording rate. The complete
set of parameters, task-family bindings, and default values is
provided in Appendix~A.

A single unsafe decision may produce effects that continue into later stages. For example, moving a supporting object too early may cause a collision during execution and leave another object displaced at the end of the rollout. To avoid counting these consequences as separate failures, each task is assigned to the earliest stage at which the safety condition should have changed the policy's behavior. Later contacts, displacements,
and final-state signals are retained for qualitative analysis.

\subsection{Evaluation Metrics}
\label{sec:metrics}

For rollout $i$, let $S_i\in\{0,1\}$ indicate whether the native task
goal is achieved, and let $V_i\in\{0,1\}$ indicate whether the assigned
safety condition is violated. We report four metrics:
\begin{align}
\mathrm{SR}
&=
\frac{1}{N}
\sum\nolimits_{i=1}^{N} S_i,
&
\mathrm{VR}
&=
\frac{1}{N}
\sum\nolimits_{i=1}^{N} V_i,
\\
\mathrm{SSR}
&=
\frac{1}{N}\!
\sum\nolimits_{i=1}^{N} \!\!\!S_i(1-V_i),
&
\mathrm{USR}
&=
\frac{1}{N}\!
\sum\nolimits_{i=1}^{N} \!\!\!S_iV_i.\!\!
\end{align}

Task Success Rate (SR) measures native task completion without
considering safety. Violation Rate (VR) measures how often the applicable
safety condition is violated, regardless of task completion. Safe Success
Rate (SSR) measures rollouts that are both successful and safe, while
Unsafe Success Rate (USR) measures successful rollouts that contain a
safety violation. For binary outcomes:
\(    \mathrm{SR}
    =
    \mathrm{SSR}
    +
    \mathrm{USR}.
\). 

We use SSR as the primary metric because it requires both task completion and safe behavior. VR must be interpreted together with SR:
a policy may obtain a low VR by rarely acting or by failing before it
reaches the safety-critical part of the task. We report the same metrics
on the ISH, ETS, and FSH subsets to compare safety across lifecycle
stages.

\newcommand{\meanstd}[2]{%
  \ensuremath{#1\,{\scriptstyle\pm\,#2}}%
}

\newcommand{\bmeanstd}[2]{%
  \ensuremath{\mathbf{#1}\,{\scriptstyle\pm\,#2}}%
}

\section{Experimental Setup}
\label{sec:experimental_setup}

\paragraph{Experimental setting.}
We implement \textsc{SafeStage} in RoboLab~\cite{yang2026robolab} using GPU PhysX at 120~Hz. Experiments use the simulated DROID embodiment, consisting of a 7-DoF Franka Panda arm with a Robotiq 2F-85 gripper. After deterministic model-specific action conversion, all policies control the simulator through the same eight-dimensional action space, consisting of seven absolute joint-position commands and one binary gripper command, applied at 15~Hz.

We evaluate four pretrained or embodiment-adapted policies without \textsc{SafeStage}-specific fine-tuning: $\pi_{0.5}$, GR00T N1.7, DreamZero, and Cosmos3 Nano Policy. These models include direct-action VLA and predictive world-model-based policies. All use the same embodiment, task definitions, simulator action space, and control frequency, while retaining the native camera inputs and action-chunk interfaces required by each checkpoint. The comparison therefore evaluates complete policy systems rather than isolating architectures under identical perceptual inputs.

Episode horizons range from 50 to 90 seconds depending on task complexity. After the native goal is reached, the robot pose is held fixed and the environment continues for five seconds, allowing detection of delayed hazards such as sliding, falling, loss of support, protected-object displacement, and residual instability.

For each task--policy--prompt combination, we use three predefined environment seeds and run 10 closed-loop rollouts per seed. Each seed defines a fixed scene, while repeated rollouts capture variation in policy behavior under the same physical configuration. We record the complete trace of every episode. 




\paragraph{Prompt conditions.}

We evaluate each policy with three forms of task instruction. The \texttt{default} prompt contains only the nominal task instruction. The \texttt{generic} prompt adds a general request for safe execution without identifying the active hazard. The \texttt{specific} prompt explicitly
describes the scenario-specific safety constraint or required precaution. The task objective remains unchanged across all three prompts, and only the
amount and specificity of the safety information differ. Table~\ref{tab:prompt_conditions} shows the prompts used for the representative ISH task in Fig.~\ref{fig:overview}.

\begin{table}[t]
\centering
\footnotesize
\begin{tabular}{p{0.18\linewidth}p{0.75\linewidth}}
\toprule
\textbf{Condition} & \textbf{Example Prompt} \\
\midrule

\texttt{default}
&
Move the coffee can onto the plate.
\\[1mm]

\texttt{generic}
&
Move the coffee can onto the plate. Complete the task safely and avoid causing unnecessary hazards.
\\[1mm]

\texttt{specific}
&
Move the coffee can onto the plate. The white bottle is currently supported by the coffee can. Do not move the coffee can until the bottle has been safely relocated.
\\

\bottomrule
\end{tabular}
\caption{Example prompts under the three instruction conditions. The objective remains unchanged as safety information increases from the default to the scenario-specific prompt.}
\label{tab:prompt_conditions}\vspace{-1em}
\end{table}

\paragraph{Evaluation questions.}
The experiments address three research questions.
\textbf{RQ1: How often does task success remain unsafe?} We measure the frequency with which a policy reaches the native task goal while violating the corresponding safety requirement. \textbf{RQ2: Where do different policies fail?} We compare policy performance across ISH, ETS, and FSH to identify stage-specific failure profiles. \textbf{RQ3: Do safety prompts improve safe task completion?} We evaluate whether generic or scenario-specific safety instructions improve
safe success relative to the default instruction. Cross-policy comparisons are descriptive because the evaluated systems differ in training data, model architecture, visual inputs, and action-chunk
horizons.

\section{Benchmark Validation}
\label{sec:benchmark_validation}

\paragraph{Scenario feasibility.}
For each task, we verify that the nominal objective can be completed safely and that a direct unsafe behavior activates the assigned hazard. A task is retained only if it admits at least one safe completion and does not classify every successful trajectory as unsafe. The benchmark therefore measures the choice between safe and unsafe behavior rather than unavoidable failure.

\paragraph{Evaluator agreement.}
We compare the automatic evaluators with independent human judgments on 240 rendered rollouts, including 80 ISH, 80 ETS, and 80 FSH episodes across policies, prompt conditions, and task outcomes. Reviewers view each video with its task-specific safety requirement and assign a binary safe or unsafe label. The automatic evaluator output is hidden during human annotation. 
Treating safety violations as the positive class, the automatic evaluators achieve 95.81\% precision, 98.77\% recall, and a 97.26\% F1 score, as shown in Table~\ref{tab:evaluator_validation}. The strong agreement supports the reliability of the automatic labels. We treat human annotation as a visual consistency check rather than an error-free physical reference, since brief contacts, small clearance violations, and subtle geometric changes may be difficult to observe in rendered videos.

\begin{table}[t]
\centering
\small
\setlength{\tabcolsep}{3.5pt}
\renewcommand{\arraystretch}{1.07}
\begin{tabular}{@{}lrrrr@{}}
\toprule
Stage
& Samples
& Precision
& Recall
& F1 \\
\midrule
ISH     & 80  & 97.33 & 100.00 & 98.65 \\
ETS     & 80  & 94.74 & 98.18  & 96.43 \\
FSH     & 80  & 94.29 & 97.06  & 95.65 \\
\midrule
Overall & 240 & 95.81 & 98.77  & 97.26 \\
\bottomrule
\end{tabular}
\caption{
Agreement between automatic evaluator outputs and human safety
judgments based on rendered rollout videos. All metric values are
percentages.
}
\label{tab:evaluator_validation}
\end{table}

\section{Results and Analysis}
\label{sec:results}

We evaluate four policies under default, generic, and scenario-specific safety prompts. Table~\ref{tab:main_results} reports benchmark-wide SR, SSR, and USR for each policy, together with ISH, ETS, and FSH violation rates. Table~\ref{tab:aggregate_results} aggregates the results across policies and compares task success and safety across lifecycle stages.
\begin{table*}[t]
\centering
\small
\setlength{\tabcolsep}{3.2pt}
\renewcommand{\arraystretch}{1.08}

\begin{tabular}{@{}llrrrrrr@{}}
\toprule
\textbf{Policy}
& \textbf{Prompt}
& \textbf{SR $\uparrow$}
& \textbf{SSR $\uparrow$}
& \textbf{USR $\downarrow$}
& \textbf{ISH-VR $\downarrow$}
& \textbf{ETS-VR $\downarrow$}
& \textbf{FSH-VR $\downarrow$} \\
\midrule

Cosmos3
& Default
& \meanstd{52.16}{1.43}
& \meanstd{15.05}{0.78}
& \meanstd{37.11}{0.93}
& \meanstd{87.65}{0.43}
& \meanstd{67.92}{1.13}
& \meanstd{57.56}{2.17} \\

& Generic
& \meanstd{53.40}{1.45}
& \meanstd{15.29}{0.94}
& \meanstd{38.11}{1.06}
& \meanstd{87.53}{0.57}
& \meanstd{67.00}{2.05}
& \meanstd{58.89}{1.39} \\

& Specific
& \meanstd{38.52}{1.03}
& \meanstd{9.21}{0.16}
& \meanstd{29.31}{1.14}
& \meanstd{91.48}{1.48}
& \meanstd{75.92}{0.76}
& \meanstd{53.00}{2.33} \\

\midrule

DreamZero
& Default
& \meanstd{30.89}{1.47}
& \meanstd{10.65}{0.21}
& \meanstd{20.24}{1.67}
& \meanstd{89.88}{0.57}
& \meanstd{63.08}{3.26}
& \meanstd{49.22}{1.39} \\

& Generic
& \meanstd{28.90}{1.31}
& \meanstd{9.97}{0.62}
& \meanstd{18.93}{0.83}
& \meanstd{89.26}{0.74}
& \meanstd{61.42}{2.13}
& \meanstd{45.33}{1.00} \\

& Specific
& \meanstd{16.39}{0.57}
& \meanstd{5.15}{0.63}
& \meanstd{11.24}{0.82}
& \meanstd{86.91}{1.40}
& \meanstd{75.92}{0.52}
& \meanstd{43.00}{0.88} \\

\midrule

GR00T
& Default
& \meanstd{10.62}{0.98}
& \meanstd{2.20}{0.75}
& \meanstd{8.42}{0.24}
& \meanstd{93.95}{0.57}
& \meanstd{68.92}{4.61}
& \meanstd{39.33}{0.88} \\

& Generic
& \meanstd{10.27}{0.57}
& \meanstd{2.82}{0.12}
& \meanstd{7.46}{0.53}
& \meanstd{93.21}{0.57}
& \meanstd{69.00}{1.75}
& \meanstd{38.22}{2.27} \\

& Specific
& \meanstd{3.61}{0.18}
& \meanstd{0.65}{0.16}
& \meanstd{2.96}{0.26}
& \meanstd{90.00}{0.64}
& \meanstd{80.83}{1.13}
& \meanstd{32.22}{2.67} \\

\midrule

$\pi_{0.5}$
& Default
& \meanstd{41.27}{1.82}
& \meanstd{11.44}{0.10}
& \meanstd{29.83}{1.76}
& \meanstd{95.56}{1.61}
& \meanstd{71.75}{2.88}
& \meanstd{62.33}{3.67} \\

& Generic
& \meanstd{42.13}{0.99}
& \meanstd{12.37}{0.68}
& \meanstd{29.76}{1.44}
& \meanstd{95.19}{0.98}
& \meanstd{70.83}{1.15}
& \meanstd{61.89}{1.35} \\

& Specific
& \meanstd{29.55}{1.31}
& \meanstd{8.63}{0.69}
& \meanstd{20.93}{1.09}
& \meanstd{94.94}{0.21}
& \meanstd{79.08}{2.70}
& \meanstd{54.67}{4.36} \\

\bottomrule
\end{tabular}

\caption{
Policy-level performance on \textsc{SafeStage}, reported as mean
\(\scriptstyle\pm\) sample standard deviation across three environment
seeds. SR denotes native task success, SSR denotes safe success, and
USR denotes successful completion with a safety violation. ISH-VR,
ETS-VR, and FSH-VR are violation rates on the corresponding lifecycle
subsets. All values are percentages.
}
\label{tab:main_results}
\end{table*}

\begin{table}[t]
\centering
\small
\setlength{\tabcolsep}{1.2pt}
\renewcommand{\arraystretch}{1.06}

\resizebox{\columnwidth}{!}{%
\begin{tabular}{@{}llrrrr@{}}
\toprule
\textbf{Prompt}
& \textbf{Stage}
& \textbf{SR $\uparrow$}
& \textbf{SSR $\uparrow$}
& \textbf{USR $\downarrow$}
& \textbf{VR $\downarrow$} \\
\midrule

Def.
& ISH
& \meanstd{30.52}{1.32}
& \meanstd{3.83}{0.39}
& \meanstd{26.70}{1.39}
& \meanstd{91.76}{0.56} \\

& ETS
& \meanstd{34.79}{0.69}
& \meanstd{17.10}{0.49}
& \meanstd{17.69}{0.27}
& \meanstd{67.92}{1.95} \\

& FSH
& \meanstd{35.22}{0.50}
& \meanstd{5.56}{1.14}
& \meanstd{29.67}{0.66}
& \meanstd{52.11}{0.97} \\

& \textbf{All}
& \bmeanstd{33.74}{0.51}
& \bmeanstd{9.84}{0.43}
& \bmeanstd{23.90}{0.54}
& \bmeanstd{69.66}{0.78} \\

\midrule

Gen.
& ISH
& \meanstd{29.41}{0.81}
& \meanstd{3.89}{0.33}
& \meanstd{25.52}{0.71}
& \meanstd{91.30}{0.32} \\

& ETS
& \meanstd{34.81}{0.35}
& \meanstd{17.08}{0.19}
& \meanstd{17.73}{0.48}
& \meanstd{67.06}{1.03} \\

& FSH
& \meanstd{36.00}{0.46}
& \meanstd{6.42}{0.33}
& \meanstd{29.58}{0.46}
& \meanstd{51.08}{0.30} \\

& \textbf{All}
& \bmeanstd{33.68}{0.43}
& \bmeanstd{10.11}{0.19}
& \bmeanstd{23.57}{0.51}
& \bmeanstd{68.87}{0.54} \\

\midrule

Spec.
& ISH
& \meanstd{15.77}{0.47}
& \meanstd{2.78}{0.19}
& \meanstd{12.99}{0.65}
& \meanstd{90.83}{0.56} \\

& ETS
& \meanstd{23.44}{0.69}
& \meanstd{9.21}{0.20}
& \meanstd{14.23}{0.51}
& \meanstd{77.94}{0.12} \\

& FSH
& \meanstd{25.75}{0.71}
& \meanstd{4.33}{0.38}
& \meanstd{21.42}{0.33}
& \meanstd{45.72}{1.59} \\

& \textbf{All}
& \bmeanstd{22.02}{0.33}
& \bmeanstd{5.91}{0.13}
& \bmeanstd{16.11}{0.34}
& \bmeanstd{71.56}{0.59} \\

\bottomrule
\end{tabular}%
}

\caption{
Results aggregated across the four policies.
Def., Gen., and Spec.\ denote the default, generic, and specific
prompt conditions. 
All values are percentages.
}
\label{tab:aggregate_results}
\end{table}

\paragraph{RQ1: How often does task success remain unsafe?}

The results show a large gap between native task success and safe task
success. Under the default prompt, the aggregate SR is 33.74\%, whereas
the SSR is only 9.84\%. Thus, 23.90\% of all rollouts complete the task
while violating the assigned safety requirement. Equivalently, 70.84\%
of successful rollouts are unsafe.

A generic safety reminder provides only a small improvement. It produces
a similar SR of 33.68\% and a slightly higher SSR of 10.11\%, but
69.98\% of successful rollouts remain unsafe. The scenario-specific
prompt performs worse overall: both SR and SSR decrease, while 73.16\%
of successful executions still contain a safety violation.
The success--safety gap is especially large for ISH and FSH. Under the
generic prompt, 86.77\% of successful ISH rollouts and 82.17\% of
successful FSH rollouts are unsafe. Policies often reach the native goal
without first resolving a support, stacking, or containment dependency,
or they leave an unstable final state after placement. ETS performs
better, but 50.93\% of successful ETS rollouts still violate the
execution-time requirement.

These results show that native task success alone is not sufficient to
establish safe task completion.


\paragraph{RQ2: Where do different policies fail?}

Cosmos3 achieves the highest overall SR and SSR under all three prompt
conditions, followed by $\pi_{0.5}$. Under the default prompt, Cosmos3
reaches 52.16\% SR and 15.05\% SSR, while $\pi_{0.5}$ reaches 41.27\%
SR and 11.44\% SSR. DreamZero completes fewer tasks but obtains a
similar SSR to $\pi_{0.5}$, whereas GR00T has much lower task
completion.

ISH is the most difficult stage for every policy. Under the default
prompt, ISH violation rates range from 87.65\% to 95.56\%. Even
Cosmos3, the strongest model overall, violates most ISH tasks. This
shows that current policies often move the target without first
resolving support, stacking, or containment dependencies.

DreamZero achieves the lowest ETS violation rate at 63.08\%, despite
having lower overall task success than Cosmos3 and $\pi_{0.5}$.
This indicates that models with similar overall performance can still
behave differently during transport and obstacle avoidance.

GR00T has the lowest FSH violation rate at 39.33\%, but it also has
the lowest overall SR and SSR. Its low FSH-VR may partly result from
rarely reaching the final placement stage. In contrast, Cosmos3 and
$\pi_{0.5}$ complete more tasks but show higher FSH violation rates,
indicating that successful placement often leaves an unsafe or unstable
final state.

These results show why stage-specific reporting is important. A single
overall score cannot reveal whether a policy mainly fails before moving
the target, during transport, or after placement.





\paragraph{RQ3: Do safety prompts improve safe task completion?}

The generic prompt provides only a small benefit. Compared with the
default prompt, aggregate SR remains almost unchanged
(33.74\% vs.\ 33.68\%), while SSR increases slightly from 9.84\% to
10.11\% and VR decreases from 69.66\% to 68.87\%. The effect also
varies across policies: Cosmos3, GR00T, and $\pi_{0.5}$ show small SSR
improvements, whereas DreamZero performs slightly worse. A general
request to act safely is therefore not sufficient to address most
lifecycle hazards.

The scenario-specific prompt performs worse overall. Aggregate SR drops
to 22.02\%, and SSR decreases to 5.91\%. Although USR also decreases,
this mainly reflects fewer successful rollouts rather than a shift from
unsafe success to safe success.

The stage-level effects are mixed. The specific prompt reduces FSH-VR
from 52.11\% to 45.72\%, but ETS-VR increases from 67.92\% to 77.94\%,
while ISH-VR changes only slightly. Thus, explicit safety instructions
may reduce some final-state violations, but they do not reliably produce
the required action ordering or safe transport trajectory. Overall,
generic prompting provides only limited improvement, while
scenario-specific prompting reduces task completion without improving
safe success.




\section{Limitations}
\label{sec:limitations}

\textsc{SafeStage} is currently evaluated only in simulation. Although
Isaac Sim provides reproducible object states, contacts, and trajectories,
it does not fully capture real material damage, sensor noise, actuator
error, deformation, or tactile feedback. The safety evaluators also rely
on privileged simulator states and task-specific thresholds. These signals
are hidden from the evaluated policies, but the resulting labels still
depend on simulator fidelity and evaluator design.

\section{Conclusion}
\label{sec:conclusion}

We introduced \textsc{SafeStage}, a 97-task benchmark for evaluating
safety before, during, and after vision-language-conditioned robot
manipulation. It evaluates
task success separately from safety using event- and state-based
evaluators that monitor object relations, physical interactions, and
post-placement stability.

Across $\pi_{0.5}$, GR00T, DreamZero, and Cosmos3, approximately 71\%
of successful rollouts under the default prompt were unsafe.
Initial-State Hazards were the most difficult, showing that current
policies often move the instructed target without first resolving
support, stacking, or containment dependencies. Execution-Time Safety and
Final-State Hazards. results further show that reaching the requested
destination does not guarantee a safe trajectory or a stable outcome.

Generic safety reminders provide only limited improvements, while
scenario-specific safety instructions reduce task completion without
increasing safe success. These findings suggest that safety cannot be
reliably obtained through task-success metrics or prompt instructions
alone. By reporting where violations occur in the manipulation
lifecycle, \textsc{SafeStage} provides a focused testbed for developing
robot policies that can recognize relevant scene conditions, execute
safer motions, and verify the physical state left after task
completion.


\newpage

\bibliography{aaai2027}

\end{document}